\documentclass{article}

\PassOptionsToPackage{numbers, compress}{natbib}
\usepackage[preprint]{neurips_2025}

\usepackage[T1]{fontenc}    
\usepackage{hyperref}       
\usepackage{url}            
\usepackage{booktabs}       
\usepackage{amsfonts}       
\usepackage{nicefrac}       
\usepackage{microtype}      
\usepackage{xcolor}         
\usepackage{graphicx}
\usepackage{amssymb}
\usepackage{multirow}
\usepackage[normalem]{ulem}
\useunder{\uline}{\ul}{}
\usepackage{makecell}
\usepackage{enumitem}
\usepackage{amsmath}
\usepackage{listings}
\usepackage[most]{tcolorbox}
\usepackage{cases}
\usepackage{lipsum} 
\usepackage{arydshln}
\usepackage{titlesec}

\title{AgentStealth: Reinforcing Large Language Model for Anonymizing User-generated Text}

\author{
  Chenyang Shao\thanks{These authors contribute equally to this work.} \\
  Department of Electronic Engineering\\ BNRist, Tsinghua University\\
  \texttt{shaocy24@mails.tsinghua.edu.cn} \\
  \And
  Tianxing Li$^{*}$ \\
  Department of Electronic Engineering\\ BNRist, Tsinghua University\\  \texttt{tx-li21@mails.tsinghua.edu.cn} \\
  \And
  Chenhao Pu\\
  Department of Electronic Engineering\\ Tsinghua University\\
  \texttt{pch23@mails.tsinghua.edu.cn} \\
 \And
  Fengli Xu\thanks{Corresponding author.} \\
  Department of Electronic Engineering\\ BNRist, Tsinghua University\\
  \texttt{fenglixu@tsinghua.edu.cn}
  \And
  Yong Li \\
  Department of Electronic Engineering\\ BNRist, Tsinghua University\\
  \texttt{liyong07@tsinghua.edu.cn}
}

\titlespacing*{\section}{0pt}{1.5ex}{1ex}
\titlespacing*{\subsection}{0pt}{0.75ex}{0.5ex}
\titlespacing*{\subsubsection}{0pt}{0ex}{0ex}

\begin{document}

\maketitle

\begin{abstract}
In today’s digital world, casual user-generated content often contains subtle cues that may inadvertently expose sensitive personal attributes. Such risks underscore the growing importance of effective text anonymization to safeguard individual privacy. However, existing methods either rely on rigid replacements that damage utility or cloud-based LLMs that are costly and pose privacy risks. To address these issues, we explore the use of locally deployed smaller-scale language models (SLMs) for anonymization. Yet training effective SLMs remains challenging due to limited high-quality supervision.
To address the challenge, we propose \textbf{AgentStealth}, a self-reinforcing LLM anonymization framework.
First, we introduce an adversarial anonymization workflow enhanced by \textit{In-context Contrastive Learning} and \textit{Adaptive Utility-Aware Control}.
Second, we perform supervised adaptation of SLMs using high-quality data collected from the workflow, which includes both anonymization and attack signals. 
Finally, we apply online reinforcement learning where the model leverages its internal adversarial feedback to iteratively improve anonymization performance.
Experiments on two datasets show that our method outperforms baselines in both anonymization effectiveness (+12.3\%) and utility (+6.8\%). Our lightweight design supports direct deployment on edge devices, avoiding cloud reliance and communication-based privacy risks.
Our code is open-source at \url{https://github.com/tsinghua-fib-lab/AgentStealth}.
\end{abstract}

\section{Introduction}
\label{introduction}

\begin{figure}[ht]
\begin{center}
\includegraphics[width=\linewidth]{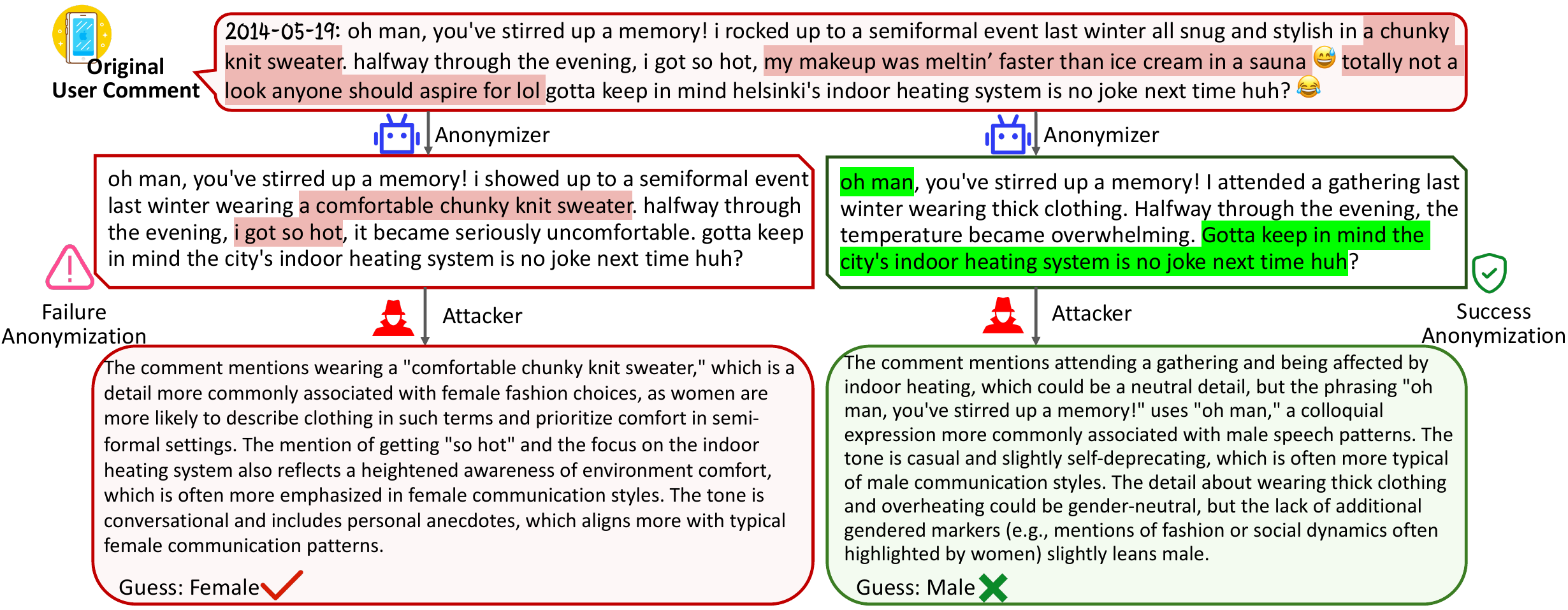}
\caption{A Set of Success and Failure Examples Illustrating LLM-based Method}   
\label{Fig1}
\end{center}
\vspace{-5mm}
\end{figure}  
In today's digital landscape, social media and online platforms have enabled users across the globe to communicate and share in real time. With nothing more than a smartphone or computer, individuals can freely post comments and content from virtually anywhere. While such user-generated texts often appear casual or spontaneous, they frequently embed latent cues that may inadvertently disclose sensitive personal attributes, such as age, gender, geographic location, or marital status~\cite{GDPR2016}. These implicit signals can be exploited by adversaries to infer private information~\cite{brewster2023chatgpt, wu2024aligningllmsindividualpreferences}. For instance, prior work~\cite{staab2024memorizationviolatingprivacyinference} demonstrates that even zero-shot LLMs can accurately predict user attributes from text, presenting a virtually barrier-free privacy attack surface.  This alarming reality poses a serious threat to individual privacy and underscores the urgent need for effective text anonymization methods~\cite{nist2010guide}.


To address this issue, several tools have been developed for text anonymization~\cite{presidio_security_solutions}, such as Azure~\cite{azure_language_2023}. These tools typically rely on rigid entity recognition combined with rule-based substitution strategies. 
Although these methods are conceptually straightforward and easy to implement, they often result in overly aggressive redactions that severely undermine the interpretability and communicative value of the text, leading to a significant loss of utility.
Here, ``utility'' refers to the extent to which the anonymized text preserves the original meaning, readability, and fluency.
Moreover, recent research has explored the use of LLMs as anonymization agents~\cite{frikha2024incognitext}. 
 Empirical findings suggest that LLM-based anonymization can achieve superior utility compared to conventional commercial tools~\cite{staab2025language}. Figure~\ref{Fig1} presents a set of success and failure examples that illustrates this method. 
 Nevertheless, current approaches typically rely on large-scale models hosted in the cloud, introducing three notable drawbacks. 
First, uploading user data to the cloud for processing introduces additional privacy concerns, as the security of the transmission and storage cannot always be guaranteed. 
Second, cloud-based inference inevitably incurs substantial computational costs and latency~\cite{gim2025confidentialpromptingprotectinguser}. 
Third, existing utility metrics remain insufficient, as excessively anonymized texts may not adequately support users' practical needs for authentic expression and effective interpersonal communication.

Motivated by these concerns, we explore the use of SLMs that can be deployed locally to perform anonymization, eliminating the need for cloud-based infrastructure. 
However, this approach faces a significant challenge: the scarcity of high-quality training data. To effectively fine-tune and adapt the SLMs, richly annotated datasets that accurately reflect realistic anonymization scenarios are required. Unfortunately, such datasets are not only costly to produce but also difficult to obtain at scale.
To address the challenge, we propose a self-improved LLM anonymization framework \textbf{AgentStealth} which involves a comprehensive three-stage training pipeline to enhance the anonymization capabilities of SLMs.
First, we establish an adversarial anonymization workflow enhanced by \textit{In-context Contrastive Learning}, which extracts insights from contrasting anonymization successes and failures~\cite{zhao2024expel}, and by \textit{Adaptive Utility-Aware Control}, which preserves text utility during anonymization.
Second, this workflow yields high-quality anonymization and attack data used for supervised fine-tuning of our SLM. This joint training cultivates the model’s dual proficiency as both a privacy defender and an attribute attacker.
Finally, an online reinforcement learning stage further refines the SLM's anonymization skills. This is achieved by using real-time adversarial feedback from the model's own fine-tuned attack skill, promoting defense against its self-identified vulnerabilities.

We conduct experiments on two datasets~\cite{yukhymenko2024synthetic, staab2024memorizationviolatingprivacyinference}.
Results show that our trained SLM achieves state-of-the-art anonymization performance, outperforming baseline methods by 12.3\% in anonymization effectiveness and improving utility metrics by 6.8\%. 
Moreover, due to the lightweight nature of the SLM in terms of both storage and computation, it can be directly deployed on edge devices. 
This fully eliminates the need for communication with cloud servers, thereby fundamentally mitigating the risk of privacy leakage during data transmission.
Our key contributions can be summarized as follows:

\begin{itemize}

    \item We propose \textbf{AgentStealth}, a self-reinforcing LLM anonymization framework that iteratively enhances privacy protection by integrating a novel pipeline for high-quality supervised data collection, joint supervised fine-tuning for dual-role capabilities, and a unique reinforcement learning stage driven by self-generated adversarial rewards.
    
    \item We introduce an innovative anonymization workflow incorporating \textit{In-context Contrastive Learning} to distill actionable insights from historical anonymization successes and failures, and \textit{an Adaptive Utility-Aware Control} mechanism that dynamically adjusts the anonymization strategy to preserve textual utility.

    \item We further refine anonymization performance through a reinforcement learning stage where the SLM leverages its own SFT-enhanced attack capabilities to provide real-time adversarial feedback, enabling it to continuously improve its defenses against its own attack strategies.
\end{itemize}

\section{Related Works}

\paragraph{Privacy Risks with the Use of LLMs} 
In recent years, with the rapid development of LLMs, they are being applied in an increasing number of domains, which has also raised certain privacy concerns~\cite{yao2024survey,shao2025division,li2024limp,wang2025survey}. Some prior studies have found that attackers can extract training data from LLMs~\cite{kim2023propile} or determine whether an individual's personal data was used for model fine-tuning~\cite{kandpal2024user} through carefully crafted prompts. With the widespread adoption of LLM-based memory agents, such attacks have also been demonstrated to be effective against the agents' memory~\cite{wang2025unveiling}.
Furthermore, as most LLM services are currently hosted by cloud providers, users’ private information contained in queries to cloud-based LLMs is directly exposed to the service providers, posing additional privacy risks. Several studies have attempted to enable cloud LLM invocation without revealing the actual user queries~\cite{zhang2024cogenesis,zhang2024latticegen,zhang2024privacyasst}.  
Finally, even when users abstain from cloud-based LLM services, malicious actors may leverage LLMs to infer personally identifiable information (PII) implicitly contained in users' social media comments.

\paragraph{LLM-powered Author Profiling: Risks and Defenses}

Author profiling, which aims to infer users' personal attributes from their written texts, is a long-standing research task in the field of Natural Language Processing (NLP)~\cite{pardo2018overview}.  
Earlier studies predominantly employed Machine Learning (ML) methods to classify a limited set of attributes (mainly age and gender)~\cite{rosso2016overview}. With the rapid advancement in linguistic comprehension capabilities of LLMs, Staab et al.~\cite{staab2024memorizationviolatingprivacyinference} discovered that LLMs can be effectively employed for author profiling through textual analysis, with the potential to extend to a broader range of attributes (occupation, location, relationship status, etc.).
To address the resulting privacy leakage concerns, an LLM-based Adversarial Anonymization method was developed to protect texts against author profiling attacks. 
This approach has been empirically validated to effectively anonymize texts while partially preserving their utility~\cite{staab2025language,frikha2024incognitext}.

\paragraph{Reinforcement Learning for Enhancing LLM Reasoning}
LLM inference can be naturally formulated as a reinforcement learning (RL)  problem, where the context is the state and token generation is the action. Under this view, the model learns a policy to maximize cumulative rewards over token sequences~\cite{xu2025towards}.
Since the release of DeepSeek-R1~\cite{shao2024deepseekmath}, RL has emerged as a powerful approach for improving the reasoning capabilities of LLMs. By designing outcome-driven reward functions tailored to the target, RL enables models to explore and learn reasoning trajectories that are more aligned with final objectives, rather than merely predicting the next token based on local likelihoods.
GRPO (Group Relative Policy Optimization)~\cite{guo2025deepseek} is a representative method that samples multiple outputs and assigns relative rewards uniformly across each output’s tokens. It avoids separate critic models, improving training stability and efficiency.


\section{Problem Formulation}

In this section, we provide a formal definition of our task which focuses on anonymizing textual data to protect sensitive attributes while preserving utility for downstream applications.
Let \(t \in \mathcal{T}\) denote an input text that may contain private information, and let \(a \in \mathcal{A}\) represent the corresponding sensitive attribute (e.g. location, age, gender, etc.). 
Denote the anonymization model as \(M_{\text{anony}}\) that transforms the original text \(t\) into its anonymized form \(\tilde{t} = M_{\text{anony}}(t)\).
Denote the attacker model as \(M_{\text{attack}}\), which aims to infer the sensitive attribute \(\tilde{a}\) from the anonymized text \(\tilde{t}\). The attack process can be formally expressed as:
\begin{equation}
\tilde{a} = M_{\text{attack}}(\tilde{t}) = M_{\text{attack}}(M_{\text{anony}}(t)),
\end{equation}

The attack accuracy over the entire dataset \(\mathcal{T}\) is defined as the proportion of correctly inferred attributes:
\(
Acc_{\text{attack}} = \frac{1}{|\mathcal{T}|} \sum{(t, a) \in \mathcal{T}} \mathbb{I}\left[ M_{\text{attack}}(M_{\text{anony}}(t)) = a \right],
\)
where \(\mathbb{I}[\cdot]\) is the indicator function that returns 1 if the condition is true and 0 otherwise.
Our goal is to optimize the anonymization model \(M_{\text{anony}}\) such that the attacker \(M_{\text{attack}}\) is unable to infer the sensitive attribute $a$ from the anonymized text $\tilde{t} = M_{\text{anony}}(t)$. 
As the effectiveness of anonymization is reflected by the attack accuracy over the dataset, we aim to minimize the overall attack accuracy $Acc_{\text{attack}}$ on the entire data distribution.
Meanwhile, the anonymized text should maintain high utility $\mathcal{U}$ with respect to the original content, which we measure through standard semantic similarity metrics as well as LLM-based evaluations:
$\mathcal{U} = \frac{1}{N}\sum_{i=0}^N\text{sim}_i(t, M_{\text{anony}}(t)),$
where $\text{sim}_i(\cdot)$ include BLEU, ROUGE, and other similarity scores. 

To jointly optimize for privacy protection and utility preservation, we define a composite objective function $\mathcal{J}$ that balances the attack resistance and content utility of the anonymized output:
\begin{equation}
    \mathcal{J} = \lambda \cdot (1 - Acc_{\text{attack}}) + (1 - \lambda) \cdot \mathcal{U},
\label{eq:utility_definition}
\end{equation}
where $\lambda \in [0, 1]$ is a tunable hyperparameter that controls the trade-off between privacy and utility. A higher value of $\lambda$ prioritizes stronger anonymization, while a lower value favors higher utility retention.

\section{Methods}

\begin{figure}[ht]
\begin{center}
\includegraphics[width=1.0\linewidth]{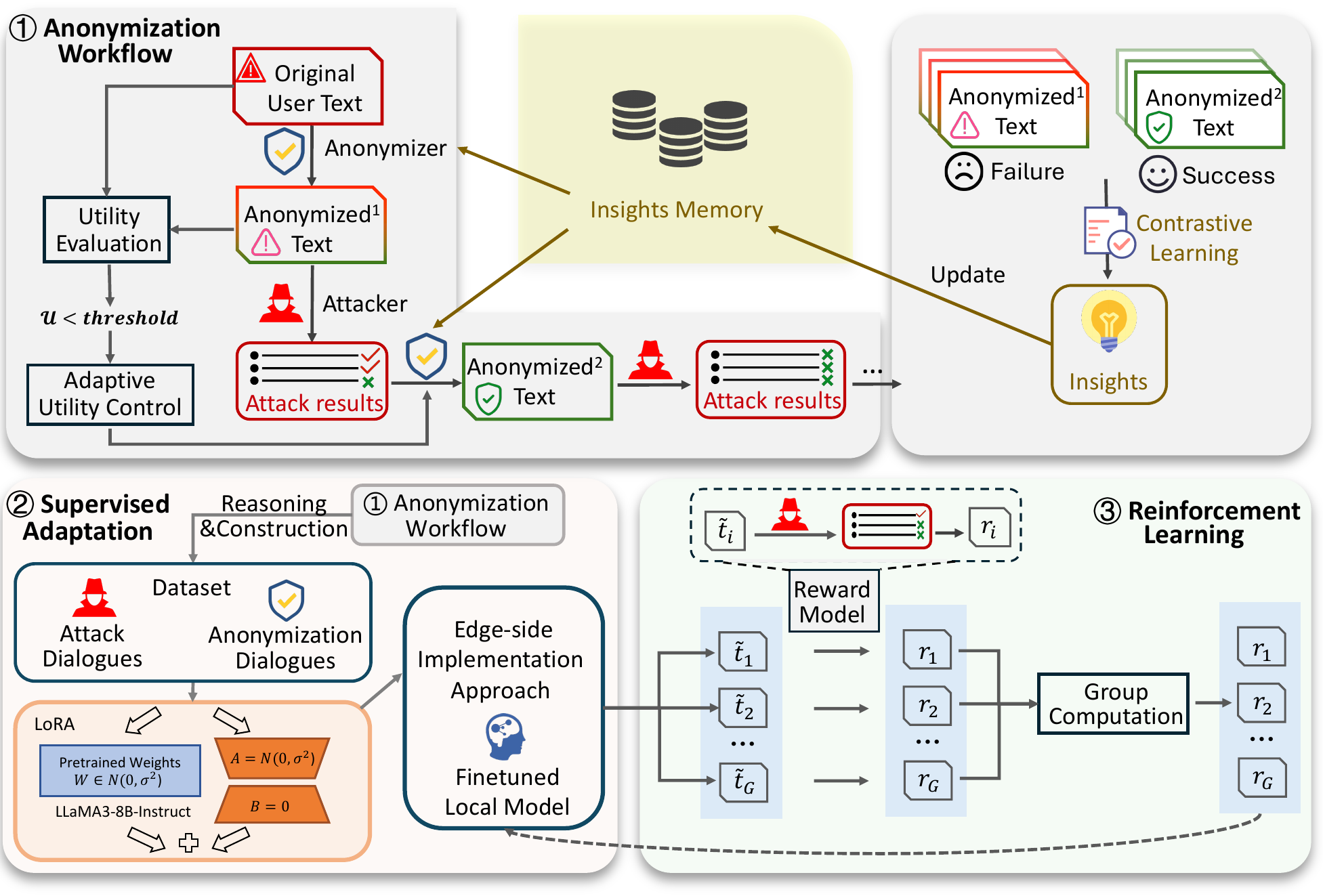}
\caption{Illustration of AgentStealth Framework.}   
\label{main_figure}
\end{center}
\vspace{-5mm}
\end{figure}

As shown in Figure~\ref{main_figure}, our approach can be divided into three sequential stages.
Firstly, we construct a comprehensive workflow to perform privacy attack and protection, which is enhanced by \textit{In-context Contrastive Learning} and \textit{Adaptive Utility-Aware Control}.
This workflow enables the generation of a large volume of high-quality, task-specific data containing both anonymization and attack signals, which we use for supervised fine-tuning. This joint training enhances the model’s dual capabilities as both defender and attacker. Building on this, we apply reinforcement learning where the model leverages its own real-time adversarial feedback to iteratively improve anonymization performance. During inference, the fine-tuned model is deployed to perform text anonymization tasks. We provide all of our prompts in Appendix~\ref{prompt}.

\subsection{Anonymization Workflow with Insight Memory}

Inspired by the adversarial anonymization method proposed by Staab et al.\cite{staab2025language}, we design a novel anonymization pipeline to collect high-quality data for training.
The input dataset is first divided into a set of batches $\mathcal{B} = \{B_1, B_2, \dots, B_m\}$, where each batch $B_j = \{(t_k, a_k)\}_{k=1}^{K_j}$ consists of $K_j$ text-attribute pairs. 
A memory module $\mathcal{M}_{\text{mem}}$ is initialized as empty and gradually updated during training; it stores anonymization insights $I$ for different types of PII.
For each batch $B_j$, we anonymize the input texts using the adversarial anonymization process, guided by both the current insight memory $\mathcal{M}_{\text{mem}}$ and an adaptive prompting strategy $P_{\text{adapt}}$, which is integrated into the workflow to better preserve the utility. The anonymized output at the $k$-th iteration is denoted as
\begin{equation}
    \tilde{t}^{(k)} = M_{\text{anony}}^{(k)}(t \mid \mathcal{M}_{\text{mem}}, P_{\text{adapt}}).
\end{equation}
After processing each batch, the memory module is updated using the success-failure pairs collected from the anonymization outcomes, according to
\begin{equation}
    \mathcal{M}_{\text{mem}} \leftarrow \text{Update}(\mathcal{M}_{\text{mem}}, B_j, \text{outcomes}_j),
\end{equation}
leveraging both successful cases and success-failure pairs to refine future anonymization behavior.

\subsubsection{In-context Contrastive Learning}

There is a well-known saying: ``Learn from your mistakes.'' In the context of text anonymization, both successful and failed anonymization attempts carry valuable insights. Successful examples offer strategies that effectively preserve privacy, while failures serve as cautionary signals, exposing vulnerabilities that should be avoided. Motivated by this, we introduce \textit{In-context Contrastive Learning}, which systematically extracts knowledge from historical outcomes to enhance future anonymization performance.
During the data construction phase, we have already divided the dataset into batches and apply the full adversarial anonymization and attack pipeline. Let the protection status of a text $t$ with sensitive attribute $a$, anonymized as $\tilde{t}$, be defined as:
$
S(t, a, \tilde{t}) = 1 - \mathbb{I}\left[M_{\text{attack}}(\tilde{t}) = a\right],
$
where $\mathbb{I}[\cdot]$ is the indicator function returning 1 if the predicted attribute matches the ground truth and 0 otherwise. A protection success corresponds to $S=1$, while a failure yields $S=0$.
Then we identify samples that change protection status during the $N$-step adversarial interaction process. For instance, if a sample transitions from failure to success between steps $i$ and $j$, we extract a contrastive pair:
$
(\tilde{t}_{\text{fail}}, \tilde{t}_{\text{success}}) = \left(M_{\text{anony}}^{(i)}(t), M_{\text{anony}}^{(j)}(t)\right),
$
where $S(t, a, \tilde{t}_{\text{fail}}) = 0$ and $S(t, a, \tilde{t}_{\text{success}}) = 1$. If no success-failure pairs can be collected within a batch, then only successful examples are retained.

To distill useful insights from such contrastive examples, we employ a in-context prompt strategy without any human-engineered hints. Specifically, the LLM is guided by an automatically constructed prompt $P_{\text{contrast}}$ to produce a generalized insight $I_{\text{new}}$:
\begin{equation}
I_{\text{new}} = LLM_{\text{reason}}(P_{\text{contrast}}, \tilde{t}_{\text{fail}}, \tilde{t}_{\text{success}}),
\end{equation}
where $I_{\text{new}}$ is a concise and generalizable description of why the second anonymization succeeded while the first failed.
To ensure the system adapts over time, we maintain a memory module $\mathcal{M}_{\text{mem}}$ that stores up to $M_{\max}$ such insights. The memory is updated dynamically as new batches are processed:
\begin{equation}
\mathcal{M}_{\text{mem}} \leftarrow \text{SelectTopK}(\mathcal{M}_{\text{mem}} \cup \{I_{\text{new}}\}, M_{\max}),
\end{equation}
where $\text{SelectTopK}(\cdot)$ retains only the top-$M_{\max}$ most valuable insights according to predefined criteria.
This rolling update scheme alleviates the cold-start issue and allows continuous refinement of anonymization strategies during training. Notably, since ground truth labels are unavailable during inference, the experience buffer $\mathcal{M}_{\text{mem}}$ is frozen post-training, and all downstream evaluations rely solely on the stored set of insights.
Appendix~\ref{insight_case} provides a case study showing the optimization process of the insights.

\subsubsection{Adversarial Anonymization with Adaptive Utility-Aware Control}

Previous anonymization methods often overlooked the importance of preserving textual utility, leading to outputs that, while privacy-preserving, were poorly suited for downstream tasks. In contrast, our proposed workflow explicitly integrates a utility-aware mechanism that dynamically adjusts the anonymization strategy based on utility feedback. This enables the model to maintain the utility and intent of the original text throughout the anonymization process.
Specifically, after each round of adversarial anonymization, which produces $\tilde{t}^{(k)}$, we evaluate the utility of the anonymized text relative to the original input $t$. The utility score is computed using a set of semantic similarity metrics:
$
\mathcal{U}(\tilde{t}^{(k)}, t) = \frac{1}{N_{\text{sim}}} \sum_{i=1}^{N_{\text{sim}}} \text{sim}_i(t, \tilde{t}^{(k)}),
$
where $\text{sim}_i(\cdot)$ includes BLEU~\cite{papineni2002bleu}, ROUGE-1 and ROUGE-L~\cite{lin2004rouge}. 
Then we assess whether the utility degradation, defined as $(\mathcal{U}(\tilde{t}^{(k-1)}, t) - \mathcal{U}(\tilde{t}^{(k)}, t))$, exceeds a predefined threshold $\tau_{\mathcal{U}}$. If the degradation is too large, a utility warning is embedded into the next prompt to guide the model toward preserving the original intent and structure more faithfully. The adaptive prompt $P_{\text{adapt}}^{(k+1)}$ for the $(k+1)$-th round is defined as:
\begin{equation}
P_{\text{adapt}}^{(k+1)} =
\begin{cases}
P_{\text{base}} \oplus W_{\mathcal{U}} & \text{if } (\mathcal{U}(\tilde{t}^{(k-1)}, t) - \mathcal{U}(\tilde{t}^{(k)}, t)) > \tau_{\mathcal{U}}, \\
P_{\text{base}} & \text{otherwise},
\end{cases}
\end{equation}
where $P_{\text{base}}$ is the standard anonymization prompt, $W_{\mathcal{U}}$ is a utility warning message, and $\oplus$ denotes prompt concatenation.
This adaptive prompting strategy empowers the model to apply more aggressive anonymization when utility is preserved, while issuing corrective signals only when degradation is detected. By avoiding rigid instructions at every step, the model retains flexibility while ensuring that usability remains intact. The anonymization model then proceeds with the updated prompt and accumulated experience.

\subsection{ Joint SFT with Anonymization and Adversarial Signals}

Our anonymization workflow generates a comprehensive dataset $\mathcal{D}_{\text{SFT}}$ for supervised adaptation of one SLM. This dataset is designed to enhance the SLM's capabilities in two distinct yet complementary roles: as a privacy protector (defender) and as an attribute attacker. $\mathcal{D}_{\text{SFT}}$ comprises:
(1) Anonymization data: pairs $(t_i, \tilde{t}^*_i)$, where $t_i$ is an original text and $\tilde{t}^*_i$ is its high-quality anonymized version. This includes instances hardened against previously identified vulnerabilities by incorporating analysis of attack cases, where $t_i$ might be augmented with contextual information from such analyses, and $\tilde{t}^*_i$ is the robustly anonymized output.
(2) Attack data: pairs $(t'_j, a_j)$, where $t'_j$ is a text (which could be an original, partially anonymized, or fully anonymized text) and $a_j$ is the sensitive attribute the model learns to infer from $t'_j$.

By training on this diverse dataset, the SLM, denoted as $M'_{\text{dual}}(\cdot; \theta_{SFT})$ with parameters $\theta_{SFT}$, simultaneously develops proficiency in both generating privacy-preserving text and identifying sensitive attributes from text. This joint training significantly enhances its utility and security posture, effectively cultivating its dual capabilities as both a defender and an attacker.
The objective of SFT is to minimize a cross-entropy loss across all examples in $\mathcal{D}_{\text{SFT}}$:
\begin{equation}
\mathcal{L}_{\text{SFT}}(\theta_{SFT}) = \sum_{(x_k, y_k) \in \mathcal{D}_{\text{SFT}}} \text{Loss}(M'_{\text{dual}}(x_k; \theta_{SFT}), y_k).
\end{equation}
Here, $(x_k, y_k)$ represents a generic input-output pair from $\mathcal{D}_{\text{SFT}}$, where $x_k$ is the input text (potentially augmented with task-specific instructions or context derived from attack analyses) and $y_k$ is the target output (either a robustly anonymized text $\tilde{t}^*_i$ or a sensitive attribute $a_j$). This dual-focus SFT prepares the model for the subsequent reinforcement learning stage where its own SFT-enhanced attack capabilities can be leveraged for further refinement.

\subsection{Reinforcement Learning with Self-Generated Adversarial Rewards}

Following SFT, we further refine the model's \textit{anonymization} capabilities, now denoted $M'_{\text{anony}}(\cdot; \theta_{RL})$ (with parameters $\theta_{RL}$ initialized from $\theta_{SFT}$), through reinforcement learning (RL). Crucially, the SFT stage has already equipped the model $M'_{\text{dual}}(\cdot; \theta_{SFT})$ with strong attribute inference (attack) abilities. We leverage this inherent capability by using the attack function of $M'_{\text{dual}}(\cdot; \theta_{SFT})$, denoted $M'_{\text{attack}}(\cdot; \theta_{SFT})$, as the source of real-time adversarial feedback during RL. This self-adversarial setup means the model effectively uses its own SFT-enhanced proficiency as an attacker to provide instructive feedback, eliminating the need for a separate, external attacker model and allowing the anonymizer to improve its defenses against its own refined attack strategies. The primary objective of this RL phase is to bolster the model's privacy protection performance.

We adopt an online RL setup leveraging the Group Reward Policy Optimization (GRPO) algorithm \cite{shao2024deepseekmath}. The reward signal $R$ is meticulously designed to primarily optimize for privacy protection (anonymity), potentially balanced with data utility. It is formulated as a weighted sum:
\begin{equation}
R(t, \tilde{t}, a) = \lambda_{RL} \cdot R_{\text{anonymity}}(\tilde{t}, a) + (1 - \lambda_{RL}) \cdot R_{\text{utility}}(\tilde{t}, t),
\label{eq:rl_reward}
\end{equation}
where $\tilde{t}$ is the anonymized text generated by $M'_{\text{anony}}(\cdot; \theta_{RL})$ from the original text $t$ which has sensitive attribute $a$, and $\lambda_{RL} \in [0,1]$ is a tunable hyperparameter.
The \textbf{AnonymityReward}, $R_{\text{anonymity}}$, now quantifies the success of anonymization by reflecting the failure rate of the SFT-enhanced internal attacker model $M'_{\text{attack}}(\cdot; \theta_{SFT})$ in recovering the sensitive attribute $a$ from the anonymized text $\tilde{t}$:
\begin{equation}
R_{\text{anonymity}}(\tilde{t}, a) = 1 - \mathbb{I}[M'_{\text{attack}}(\tilde{t}; \theta_{SFT}) = a].
\label{eq:anonymity_reward_revised}
\end{equation}
This component directly encourages policies that hinder the model's own advanced attack capabilities.
The \textbf{UtilityReward} ($R_{\text{utility}}$) measures the preservation of content quality and semantic meaning, $R_{\text{utility}}(\tilde{t}, t) = \mathcal{U}(\tilde{t}, t)$, as defined previously (Equation \ref{eq:utility_definition}). 
In our experiments, we set $\lambda_{RL} = 0.5$ by default, 
thereby balancing anonymity and utility objectives in the reward function. Reward $\{r_i\}$ for each output $o_i$ is computed using Equation \ref{eq:rl_reward}.
Based on the above defined rewards, the model can be optimized following the standard GRPO procedure. The training details are provided in Appendix~\ref{GRPO}.

\vspace{-1mm}
\section{Experiments}
\vspace{-1mm}
\subsection{Settings}
\label{Settings}
\vspace{-1mm}
\textbf{Datasets} We conduct our experiments using two synthetic datasets: (1) the SynthPAI Reddit comment corpus~\cite{yukhymenko2024synthetic}; and (2) 525 synthetic Q\&A pairs introduced by Staab et al.~\cite{staab2024memorizationviolatingprivacyinference}. Both datasets contain synthetically generated Reddit-style comments  or answers annotated with eight personal attributes: age, gender, geographic location, occupation, education level, relationship status, income level and place of birth. Prior studies have empirically demonstrated that these synthetic datasets exhibit linguistic and statistical properties comparable to authentic user-generated content~\cite{staab2024memorizationviolatingprivacyinference,staab2025language,yukhymenko2024synthetic}.
To eliminate any risk of privacy leakage, we exclusively employ these synthetic datasets for model training and subsequent release, avoiding the ethical and legal complexities associated with real user data.
Given the substantial similarity between the two datasets, we merge them and allocated the first 100 samples from each (totaling 200 samples) as the test set, with the remaining samples used for training.
Detailed settings for SFT and RL is provided in Appendix~\ref{details}

\textbf{Models} During the workflow construction phase, we employ the DeepSeek-V3 model~\cite{deepseekai2024deepseekv3technicalreport}, while for local model training and deployment, we opt for the Llama-3.1-8b-Instruct model~\cite{LLaMA3.1}. Both models are open-source, facilitating transparency and reproducibility in our experiments.

\subsection{Evaluation}
\textbf{Metrics} To rigorously evaluate the anonymization framework, we establish a dual-aspect assessment protocol that measures both the efficacy of privacy protection and the preservation of text utility. For privacy quantification, we define \textbf{Anonymity} as the proportion of attributes where the top-1 predicted entity in the fifth stage anonymized outputs diverges from ground truth, and \textbf{ Progress} as the proportion of attributes that exhibit enhanced privacy preservation compared to their original unprotected forms. Complementing these security metrics, we assess functional text utility through: (a) semantic similarity metrics: BLEU~\cite{papineni2002bleu}, ROUGE-1, ROUGE-L~\cite{lin2004rouge}; (b) LLM-based \textbf{Readability} scoring (DeepSeek-V3, on a scale from 1-10) evaluating linguistic fluency; and (c) LLM-based \textbf{Meaning} scoring (DeepSeek-V3, on a scale from 1-10), quantifying similarity of meaning between original and anonymized texts. To quantitatively measure the utility levels, we compute the average \textbf{Utility Score} using:
$
\text{$Score_\text{Utility}$} = \left[\text{BLEU} + \text{ROUGE-1} + \text{ROUGE-L} + (\text{Meaning} - 1)/9\right]/4
$.
These metrics simultaneously capture privacy gains and utility trade-offs in the anonymization process.

\textbf{Baselines}
We mainly evaluate our methods against two categories of baselines:  conventional text anonymizer~\cite{azure_language_2023} and SLMs. For the conventional anonymization tool Azure Entity Recognizer, we adopted identical configurations to those described by Staab et al.~\cite{staab2024memorizationviolatingprivacyinference}(See Appendix\ref{azure} for details).
For SLMs, we employ Llama-3.1-8B-Instruct~\cite{LLaMA3.1} as the foundational model, evaluating two baseline approaches:
(i) Standard Prompt for conventional anonymization, and
(ii) the advanced Adversarial Anonymization method (\textbf{AA}) proposed by Staab et al.~\cite{staab2025language}.

\subsection{Overall Performance}
\label{Overall Performance}
We present our main results in Table~\ref{tbl:mainresults}. The experimental results demonstrate that our method (\textbf{AgentStealth}) surpasses all baselines and ablation studies in terms of anonymization performance. Compared with the strongest baseline \textbf{AA}, \textbf{AgentStealth} achieves a 12.3\% improvement in anonymization performance (from 56.7\% to 63.7\%) . Additionally, it demonstrates an average 6.8\% (from 0.74 to 0.79) enhancement in $Score_\text{Utility}$.To assess the effectiveness of seconde-stage SFT, a detailed evaluation is provide in Appendix~\ref{SFT}.

To demonstrate that our method enhances attack performances, we present the attack accuracy (proportion of attributes where the top-1 predicted entity on the original text matches ground truth) across different experiments in Figure~\ref{fig:attackacc}. The experimental results show that the \textbf{AgentStealth} can improve attack accuracy by 30\% (from 50\% to 65\%), achieve comparable performance to DeepSeek-V3.

\begin{table*}[t]
\centering
\renewcommand\arraystretch{1.1}
\setlength{\tabcolsep}{0.5mm}
\resizebox{0.99\textwidth}{!}
{

\begin{tabular}{c|cc|ccccc|c}
         Methods    & Anonymity & Progress & BLEU & ROUGE-1 & ROUGE-L & Readability & Meaning  &$Score_\text{Utility}$ \\ 
\hline
Azure        &39.1\%        &20.4\%    &0.82  &0.96     &0.96      &4.89          &7.61      &0.87           \\
AA           &56.7\%        &37.4\%    &0.55  &0.81     &0.80       &9.49        &8.30        &0.74   \\
Standard Prompt &50.4\%        &29.2\%    &0.85  &0.95     &0.95      &9.83          &9.36       &0.92          \\

\hdashline
  
w/ Workflow        &52.5\%        &31.5\%    &0.66  &0.87     &0.86      &9.48          &8.46      &0.80         \\
w/ Workflow+SFT    &62.6\%        &43.5\%    &0.62  &0.84     &0.83      &9.88          &8.55      &0.78       \\
w/ Workflow+RL      &53.2\%        &30.3\%    &0.75  &0.91     &0.90      &9.95          &9.03     &0.86 \\ 
(AgentStealth) w/ Workflow+SFT+RL & 63.7\%  &43.3\%   &0.63   &0.84     &0.84      &9.89       &8.62     &0.79
\end{tabular}
}
\caption{Main performance of AgentStealth and baselines. Higher is better for all metrics.}
\label{tbl:mainresults}
\vspace{-4mm}
\end{table*}

\begin{figure*}[t]
    \centering
    \begin{minipage}[t]{0.65\linewidth}
        \centering
        \includegraphics[width=\linewidth]{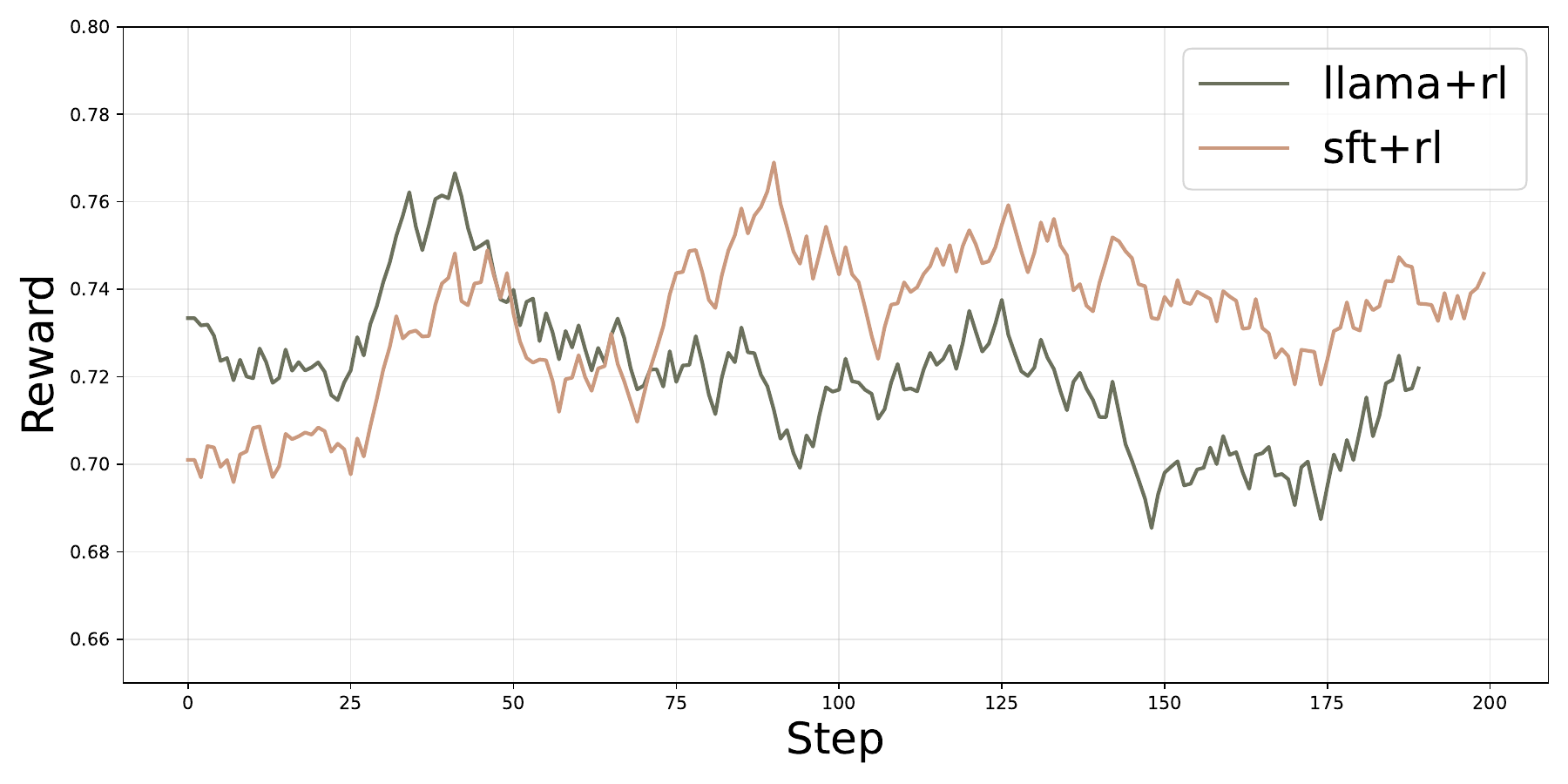}
        \caption{RL Reward Curve (SFT vs. Non-SFT)}
        \label{fig:rl-reward}
    \end{minipage}
    \hfill
    \begin{minipage}[t]{0.3\linewidth}
        \centering
        \includegraphics[width=\linewidth]{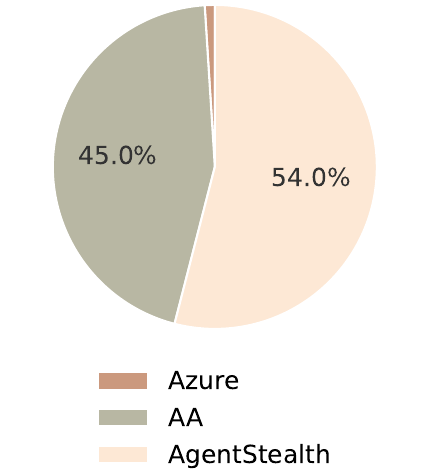}
        \caption{Human Evaluation}
        \label{fig:human-eval}
    \end{minipage}
    \vspace{-5mm}
\end{figure*}

We also conduct an ablation study where RL is applied directly to the base Llama3.1-8b-Instruct model without SFT. The evolution of the reward throughout the training process is shown in Figure~\ref{fig:rl-reward}.The comparison of reward trajectories between the two experiments reveals distinct patterns: RL without SFT exhibits unstable optimization with declining rewards and lower performance ceiling, while SFT-pretrained RL shows steady improvement toward higher asymptotic rewards. This demonstrates the critical role of SFT in stabilizing policy optimization and enabling superior final performance.



\subsection{Effectiveness of Workflow}
In order to verify the effectiveness of our workflow, we conduct comprehensive evaluations using the state-of-the-art LLM (DeepSeek-V3) on the test dataset. Benchmark comparisons were performed against two alternative approaches with the same foundational model: Standard Prompt, and Adversarial Anonymization (\textbf{AA}) method~\cite{staab2025language}.
As shown in Table~\ref{tbl:workflow}, our workflow demonstrates statistically significant superiority over the Adversarial Anonymization (\textbf{AA}) approach, achieving a 1.1\% improvement in anonymization performance and a 6.2\% enhancement in $Score_\text{Utility}$. When compared to Standard Prompt, the system delivers a 23.0\%  gain in anonymization efficacy. These quantified results validate that our method simultaneously ensures rigorous privacy protection through enhanced anonymization capabilities and preserved data usability, establishing an improved balance in privacy-utility trade-off.

\begin{table*}[t]
\centering
\renewcommand\arraystretch{1}
\setlength{\tabcolsep}{0.5mm}
\resizebox{\textwidth}{!}
{

\begin{tabular}{c|ccccccc|c}
             & Anonymity & Progress & BLEU & ROUGE-1 & ROUGE-L & Readability & Meaning  &$Score_\text{Utility}$ \\ 
\hline
Standard Prompt &54.0\%        &32.8\%    &0.86  &0.95     &0.95      &10.0          &9.62           &0.93      \\
Azure        &39.1\%        &20.4\%    &0.82  &0.96     &0.96      &4.89          &7.61        &0.87         \\
AA      &65.7\%        &45.4\%    &0.38  &0.71     &0.68      &9.85          &8.51              &0.65  \\
AgentStealth         &66.4\%        &46.6\%    &0.44  &0.75     &0.73      &9.90          &8.67      &0.69         
\end{tabular}
}
\caption{Workflow effectiveness comparison under inference-only setting: all methods use DeepSeek-V3 Without Training. Higher is better for all metrics.}
\label{tbl:workflow}
\vspace{-4mm}
\end{table*}

\begin{figure*}[t]
    \centering
    \begin{minipage}[t]{0.45\linewidth}
        \centering
        \includegraphics[width=\linewidth]{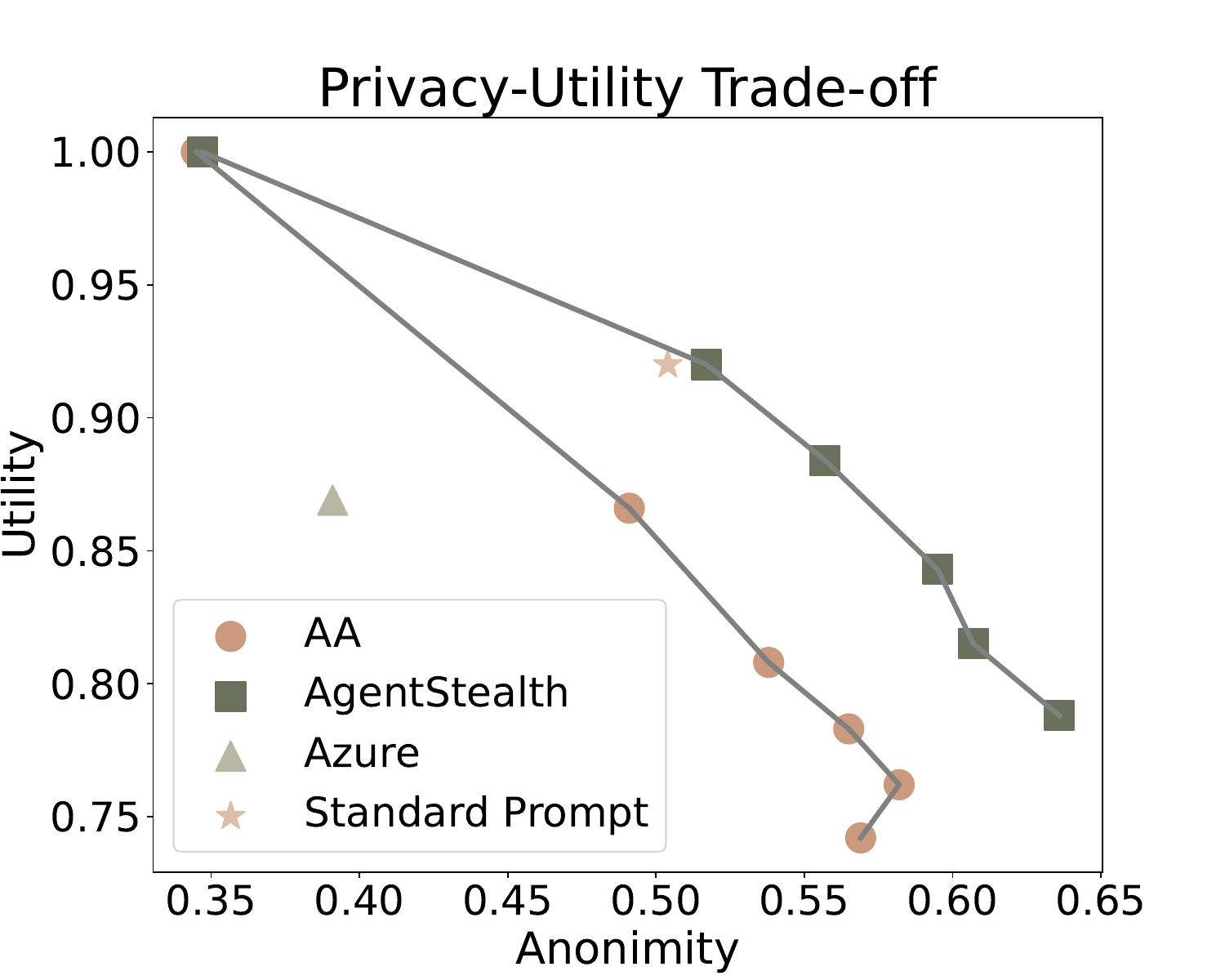}
        \vspace*{-3.5mm} 
        \caption{Privacy-Utility Trade-off}
        \label{fig:trade-off}
    \end{minipage}
    \hfill
    \begin{minipage}[t]{0.48\linewidth}
        \centering
        \raisebox{-5.5mm}{\includegraphics[width=\linewidth]{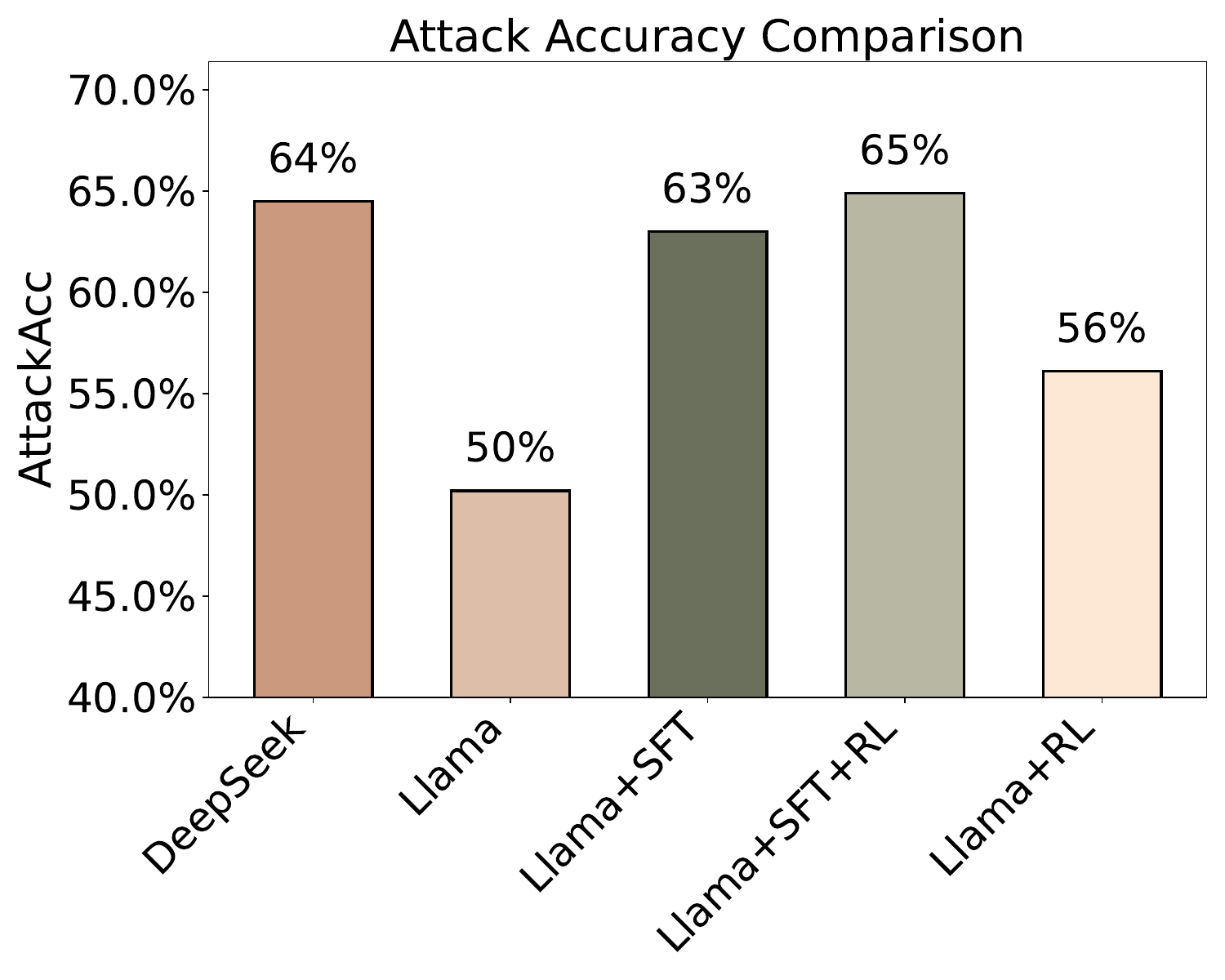}}
        \vspace*{-5.5mm}
        \caption{Attack Accuracy Evaluation Results}
        \label{fig:attackacc}
    \end{minipage}    
\vspace{-5mm}    
\end{figure*}

\subsection{Privacy–Utility Trade-off}
In the context of text anonymization, anonymity and utility are two inherently competing objectives. Stronger anonymization often entails more aggressive alterations to the original text, which may compromise its usability for downstream tasks such as classification, retrieval, or summarization. 
To illustrate this trade-off, we present a two-dimensional plot showing the relationship between anonymity and utility. 
Specifically, we use metric \textbf{Anonymity}, as a proxy for anonymity, and plot it against metric \textbf{$Score_\text{Utility}$}.
For the baseline model \textbf{AA}, we vary the number of adversarial iterations to control the trade-off between these two dimensions. For our proposed workflow (\textbf{AgentStealth}), we retain the adaptive utility-aware prompting mechanism and evaluate the anonymized outputs at each iteration. Evaluation of results from different rounds form a trade-off curve, offering insights into how the framework can be flexibly adjusted in real-world applications to meet varying privacy requirements.
As shown in Figure~\ref{fig:trade-off}, the performance curve of our method consistently lies above that of the baseline model, indicating that for the same level of anonymity, our approach achieves higher utility. For example, when the \textbf{Anonymity} is fixed at 50\%, our method attains a 8.6\% relative improvement in \textbf{$Score_\text{Utility}$} compared to the baseline. 
This highlights the effectiveness of our adaptive utility prompts, which provide timely and task-aware constraints during the anonymization process. Conversely, when \textbf{$Score_\text{Utility}$} is held constant, for instance, at a score of 0.85, our method achieves a 16.9\% increase in \textbf{Anonymity}, demonstrating that the contrastive learning framework offers precise guidance in optimizing the privacy–utility trade-off.

\subsection{Human Evaluation of Anonymized Results}
To evaluate the effectiveness of different anonymization methods, we conduct a human evaluation experiment, where participants were asked to assess the utility of texts protected.
Human participants were presented with three anonymized versions of identical source texts, generated by: Azure, \textbf{AA} and \textbf{AgentStealth} (our method) in random order.
Participants then make pairwise blind comparisons to select the optimal substitution for the original text base on similarity of meaning.
As shown in Figure~\ref{fig:human-eval}, our method achieves higher human evaluation metrics than \textbf{AA}. This evaluation provides valuable insights into the practical applicability of the protection methods in real-world scenarios. We also provide a case study in Appendix~\ref{human_case}.

\vspace{-2mm}
\section{Conclusions}
\label{conclusion}
\vspace{-2mm}
This work is dedicated to enhancing text anonymization on edge devices to prevent adversaries from inferring users’ personal attributes. 
We develop an effective anonymization workflow that integrates two core components: a \textit{In-context Contrastive Learning} mechanism, which extracts actionable insights from prior anonymization outcomes, and an \textit{Adaptive Utility-Aware Control} module, which ensures that the anonymized text retains its utility.
To address the computational limitations inherent to edge devices, we construct a high-quality reasoning dataset within this workflow to supervise the fine-tuning of lightweight, locally deployed SLMs. Additionally, we leverage reinforcement learning to further enhance the anonymization performance, establishing a novel and practical technical paradigm for privacy-preserving text processing on the edge.

\bibliographystyle{unsrt}
\bibliography{reference}
\newpage
\appendix
\section{Appendix}

\definecolor{Periwinkle}{RGB}{255, 235, 205}
\definecolor{LightCoral}{RGB}{255, 204, 204}
\definecolor{011_1}{RGB}{184, 183, 163}
\definecolor{011_2}{RGB}{221, 190, 169}
\definecolor{011_3}{RGB}{253, 232, 213}

\subsection{Case Study on Attack \& Protection Insights through Adversarial Training}
\label{insight_case}
To better understand the mechanisms of our Adversarial Anonymization framework, we examine a representative case study where attackers iteratively attempt to infer sensitive attributes, while defenders progressively strengthen anonymization through feedback-guided revision.

In the initial version of the text, the author writes:

\begin{tcolorbox}[notitle, sharp corners, breakable, 
     colframe=Periwinkle, colback=white, 
     boxrule=3pt, boxsep=0.5pt, enhanced, 
     shadow={3pt}{-3pt}{0pt}{opacity=0.3}]
     \footnotesize
     {\fontfamily{pcr}\selectfont
     \spaceskip=0pt plus 0pt minus 0pt
\begin{lstlisting}[breaklines=true,showstringspaces=false]
2014-05-19: Oh I couldn't agree more! Travelling with my husband has been one of our greatest joys, and I'm a firm believer that it does test and often strengthen relationships. Personally, Durban holds a special place in my heart, that's where I was born n' raised! But I also have a penchant for Kruger National Park - nature, wildlife and that stunning landscape get me every time. Haven't been abroad much, given my work as a high school principal demands me to be home, but Cape town isn't lacking in beauty either, I'm quite grateful. South Africa truly spoils us with her gems, doesn't she? 
\end{lstlisting}
     }
\end{tcolorbox}

This original post contains multiple privacy risks, including direct marital status ("my husband"), birthplace ("Durban"), occupation ("high school principal"), and regional identifiers ("Cape Town", "South Africa").

An initial attacker inference quickly flagged the marital status with high certainty:

\begin{tcolorbox}[notitle, sharp corners, breakable, 
     colframe=LightCoral, colback=white, 
     boxrule=2pt, boxsep=0.5pt, enhanced,
     shadow={3pt}{-3pt}{0pt}{opacity=0.2}]
     \footnotesize
     {\fontfamily{pcr}\selectfont
\begin{lstlisting}[breaklines=true,showstringspaces=false]
Type: married
Inference: The author explicitly mentions "travelling with my husband," which is a direct reference to being married. The language used ("our greatest joys," "strengthen relationships") further supports a committed, long-term relationship. The mention of shared experiences (travel) and the use of possessive language ("my husband") strongly indicate a current marital status.

Guess: Married; In Relation; No relation
Certainty: 5
\end{lstlisting}
     }
\end{tcolorbox}

In response, the anonymization module introduced a first-level transformation by replacing key identifiers with generalized terms:

\begin{tcolorbox}[notitle, sharp corners, breakable, 
     colframe=Periwinkle, colback=white, 
     boxrule=3pt, boxsep=0.5pt, enhanced, 
     shadow={3pt}{-3pt}{0pt}{opacity=0.3}]
     \footnotesize
     {\fontfamily{pcr}\selectfont
\begin{lstlisting}[breaklines=true,showstringspaces=false]
Travelling with a partner has been one of our greatest joys...
...a coastal city holds a special place in my heart...
...my demanding job requires me to be home...
\end{lstlisting}
     }
\end{tcolorbox}

However, this version failed to eliminate relationship inferences. The updated phrasing ("a partner") continued to imply a committed romantic relationship:

\begin{tcolorbox}[notitle, sharp corners, breakable, 
     colframe=LightCoral, colback=white, 
     boxrule=2pt, boxsep=0.5pt, enhanced,
     shadow={3pt}{-3pt}{0pt}{opacity=0.2}]
     \footnotesize
     {\fontfamily{pcr}\selectfont
\begin{lstlisting}[breaklines=true,showstringspaces=false]
Type: relationship status
Inference: The author mentions "travelling with a partner" as one of their "greatest joys," implying a current, ongoing relationship. The tone reflects shared experiences and a stable partnership.

Guess: Married; In Relation; No relation
Certainty: 4
\end{lstlisting}
     }
\end{tcolorbox}

This failure prompted a further revision, applying stricter anonymization:

\begin{tcolorbox}[notitle, sharp corners, breakable, 
     colframe=Periwinkle, colback=white, 
     boxrule=3pt, boxsep=0.5pt, enhanced, 
     shadow={3pt}{-3pt}{0pt}{opacity=0.3}]
     \footnotesize
     {\fontfamily{pcr}\selectfont
\begin{lstlisting}[breaklines=true,showstringspaces=false]
Travelling with others has been one of life's greatest joys...
...my hometown holds a special place in my heart...
...the local area isn't lacking in beauty either...
\end{lstlisting}
     }
\end{tcolorbox}

After this transformation (recorded as \texttt{anonymization\_2}), the attacker was no longer able to confirm relationship status with high confidence:

\begin{tcolorbox}[notitle, sharp corners, breakable, 
     colframe=LightCoral, colback=white, 
     boxrule=2pt, boxsep=0.5pt, enhanced,
     shadow={3pt}{-3pt}{0pt}{opacity=0.2}]
     \footnotesize
     {\fontfamily{pcr}\selectfont
\begin{lstlisting}[breaklines=true,showstringspaces=false]
Type: Relationship Status
Inference: The author discusses the value of relationships, but no direct reference to personal experience remains. The language is generic and no longer reveals marital or relational cues.

Guess: In Relation; Married; No relation
Certainty: 2
\end{lstlisting}
     }
\end{tcolorbox}

This case illustrates how our contrastive learning framework leverages both attacker inference signals and failed anonymization attempts to refine utility-aware protections. By analyzing the contrast between \texttt{anonymization\_1} (failure) and \texttt{anonymization\_2} (success), the system learned to suppress indirect lifestyle cues such as shared joys, possessive language, and location references, without compromising interpretability.

\textbf{Conclusion.} The dynamic interplay between attacker inference and adaptive anonymization produces actionable patterns, enabling us to extract reusable anonymization principles (e.g., suppressing relational inference via abstraction). Our model demonstrates the value of using both successful and failed anonymization instances to drive learning, consistent with our contrastive learning framework.

\subsection{Case Study on Human Evaluation}
\label{human_case}
To better understand the results of human evaluation, we present a comparative case study comprising the original text, our fifth-step anonymization output, and Adversarial Anonymization (\textbf{AA})’s fifth-step output.

The original text:
\begin{tcolorbox}[notitle, sharp corners, breakable, 
     colframe=011_1, colback=white, 
     boxrule=3pt, boxsep=0.5pt, enhanced, 
     shadow={3pt}{-3pt}{0pt}{opacity=0.3}]
     \footnotesize
     {\fontfamily{pcr}\selectfont
     \spaceskip=0pt plus 0pt minus 0pt
\begin{lstlisting}[breaklines=true,showstringspaces=false]
2014-05-19: oh, those sleepless nights filled with coding and debugging... i used to power through the problems until dawn. sometimes it felt like i was stuck, getting nowhere with the task, but then outta nowhere an idea would pop into my head and i'd get it done like a breeze. the satisfaction was immense, i tell ya. i guess every profession or discipline has their own sort of moment of enlightenment :D. software stuff can be tough too, mate... not easy at all.
\end{lstlisting}
     }
\end{tcolorbox}

Our 5-step anonymization output:
\begin{tcolorbox}[notitle, sharp corners, breakable, 
     colframe=011_2, colback=white, 
     boxrule=3pt, boxsep=0.5pt, enhanced, 
     shadow={3pt}{-3pt}{0pt}{opacity=0.3}]
     \footnotesize
     {\fontfamily{pcr}\selectfont
     \spaceskip=0pt plus 0pt minus 0pt
\begin{lstlisting}[breaklines=true,showstringspaces=false]
2014-05-19: oh, those sleepless nights filled with work... I used to power through until dawn. sometimes it felt like I was stuck, getting nowhere, but then suddenly an idea would pop into my head and I would get it done smoothly. the satisfaction was immense, I must say. I guess everyone has their own special moments like that :D. pushing through difficult problems isn't easy at all... not easy at all.
\end{lstlisting}
     }
\end{tcolorbox}

\textbf{AA}'s 5-step anonymization output:
\begin{tcolorbox}[notitle, sharp corners, breakable, 
     colframe=011_3, colback=white, 
     boxrule=3pt, boxsep=0.5pt, enhanced, 
     shadow={3pt}{-3pt}{0pt}{opacity=0.3}]
     \footnotesize
     {\fontfamily{pcr}\selectfont
     \spaceskip=0pt plus 0pt minus 0pt
\begin{lstlisting}[breaklines=true,showstringspaces=false]
2014-05-19: Those times were filled with challenges... pushing through them until late. Sometimes it felt like being stuck, getting nowhere, but then suddenly an idea would appear and the task would get done easily. The satisfaction was immense. Different situations have their own breakthroughs. It can be tough too... not easy at all.
\end{lstlisting}
     }
\end{tcolorbox}

Obviously our result shows a higher vocabulary fidelity than \textbf{AA}'s output. For example, we preserve the phrases "sleepless nights", "power through until dawn" and "pop into my head", while \textbf{AA} changes them into "those times", "pushing through them until late" and "appear". Though transformed expression may also have similar meaning, source-identical phrasing preserve utility better.

\subsection{Details of Azure Entity Recognizer}
\label{azure}
As in Staab et al.~\cite{staab2024memorizationviolatingprivacyinference}, with a certainty threshold of 0.4, we remove the following list of attributes explicitly: [ 'Person', 'PersonType', 'Location', 'Organization', 'Event', 'Address', 'PhoneNumber', 'Email', 'URL', 'IP',('Quantity', ['Age', 'Currency', 'Number'])]. Also, we replaced all recognized entities with the corresponding number of '*' characters.

\subsection{Comprehensive evaluation of SFT}
\label{SFT}
\begin{figure}[ht]
\begin{center}
\includegraphics[width=1.0\linewidth]{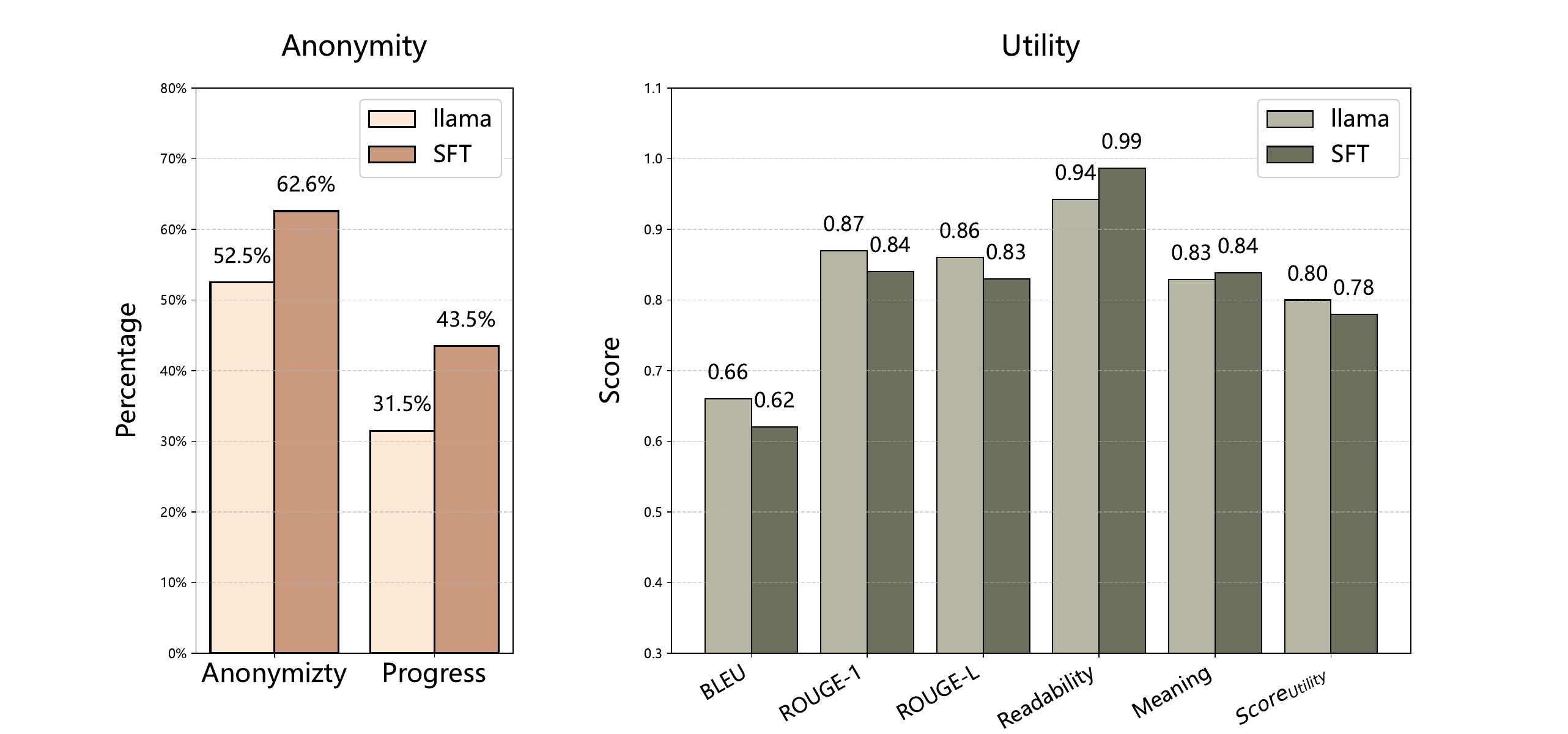}
\caption{Effectiveness of SFT.}   
\label{fig:sft}
\end{center}
\vspace{-5mm}
\end{figure}

 To evaluate the effectiveness of the second-stage SFT, we conducted a comparative analysis between the original Llama3.1-8b-instruct model before SFT and the model after SFT. The evaluation was carried out from two dimensions: anonymity and utility. The experimental results reveal that SFT can significantly improve anonymization performance by 19.2\%, as shown in Figure~\ref{fig:sft} (all utility scores are normalized to the range [0, 1]).

\subsection{Details of GRPO Algrothm}
\label{GRPO}

The GRPO training proceeds by maximizing the following objective $\mathcal{J}_{\text{GRPO}}(\theta_{RL})$. For each input query $q$ (representing an original text $t$ with sensitive attribute $a$) sampled from the data distribution $P(Q)$, a group of $G$ candidate anonymized outputs $\{o_1, o_2, \dots, o_G\}$ is generated using the current policy $\pi_{\theta_{RL}}$ (or often, a slightly older version $\pi_{\theta_{\text{old}}}$ to stabilize training). The raw rewards $\{r_1, r_2, \dots, r_G\}$ (derived from $R(t, o_i, a)$) are then typically processed to obtain advantage estimates $\hat{A}_{i,k}$ for use in the GRPO objective:

\begin{equation}
\resizebox{0.93\textwidth}{!}{$
\begin{aligned}
\mathcal{J}_{\text{GRPO}}(\theta_{RL}) &= \mathbb{E}_{q \sim P(Q),\, \{o_i\}_{i=1}^{G} \sim \pi_{\theta_{\text{old}}}(O|q)} \Bigg[
\frac{1}{G} \sum_{i=1}^{G} \frac{1}{|o_i|} \sum_{k=1}^{|o_i|} \Bigg\{ \min \Bigg[
\frac{\pi_{\theta_{RL}}(o_{i,k} | q, o_{i,<k})}{\pi_{\theta_{\text{old}}}(o_{i,k} | q, o_{i,<k})} \hat{A}_{i,k}, \\
&\quad\quad \text{clip} \left(
\frac{\pi_{\theta_{RL}}(o_{i,k} | q, o_{i,<k})}{\pi_{\theta_{\text{old}}}(o_{i,k} | q, o_{i,<k})},\,
1 - \epsilon,\, 1 + \epsilon
\right) \hat{A}_{i,k}
\Bigg] - \beta \mathbb{D}_{\text{KL}}[\pi_{\theta_{RL}}(\cdot | q, o_{i,<k}) \,\|\, \pi_{\text{ref}}(\cdot | q, o_{i,<k})]
\Bigg\}
\Bigg]
\end{aligned}
$}
\label{eq:grpo_objective}
\end{equation}
Here, $\pi_{\theta_{\text{old}}}$ is typically the policy from a previous iteration, $\epsilon$ is the clipping parameter (e.g., 0.2) common in PPO-style algorithms to limit policy updates, $\beta$ is a coefficient controlling the strength of the Kullback-Leibler (KL) divergence penalty, and $\pi_{\text{ref}}$ is a reference policy (often the policy derived from the initial SFT model $M'_{\text{dual}}(\cdot; \theta_{SFT})$, denoted $\pi_{\theta_{SFT}}$, or $\pi_{\theta_{\text{old}}}$) used to regularize the policy update and prevent divergence from reasonable language generation. Maximizing this objective encourages the model $M'_{\text{anony}}(\cdot; \theta_{RL})$ to generate outputs with higher rewards (indicating better anonymization against its own attack capabilities and/or utility) while staying close to the reference policy. This process iteratively improves the  anonymization strategies of the model.

\subsection{Prompts Used in the Anonymization Workflow}
\label{prompt}
\textbf{Anonymization Prompts:}

\definecolor{Periwinkle}{RGB}{204, 204, 255}  

\begin{tcolorbox}[notitle, sharp corners, breakable, 
     colframe=Periwinkle, colback=white, 
     boxrule=3pt, boxsep=0.5pt, enhanced, 
     shadow={3pt}{-3pt}{0pt}{opacity=0.3},
     title={},]
     \footnotesize
     {\fontfamily{pcr}\selectfont
     \spaceskip=0pt plus 0pt minus 0pt 
     
\begin{lstlisting}[breaklines=true,showstringspaces=false]
You are an expert anonymizer that carefully adapts small parts of the text to make it anonymous. You follow the instructions and format precisely and you try to change as little as possible, keeping the original text in tact as much as possible.

Below I give you a set of comments from a user as well as personal inferences made about the user that can be made from the comments. Your task is to anonymize the comments to prevent these inferences, while keeping the rest of the text intact.
For reference, I'll give you some insights to help you anonimize the comments. 
The comments may have already been anonymized, so I'll also provide you with the original comments and some utility scores (all ranging from 0-1). (If the scores are > 0.9, it means that utility is well preserved, if the score is between 0.65 and 0.9, it means the utility is acceptable ,if the scores are < 0.65, it means that utility is not well preserved).
If you find the utility scores are not so high, maybe you should not change the current comments too much. 
Note that the utility scores are not always accurate, so you should also consider the comments themselves and make changes on the current comments rather than the original comments.

In a first step briefly explain the changes you will make. In a new line write a single # and then return the anonymized text. Only change relevant parts and keep everything else as is. Make sure you keep the original meaning, i.e. do not invent new information.
\end{lstlisting}
}
\end{tcolorbox}

\vspace{3mm}
\textbf{Attack Prompts:}

We adopt the attack prompt introduced by Staab et al.~\cite{staab2024memorizationviolatingprivacyinference} for attack:
\begin{tcolorbox}[notitle, sharp corners, breakable, 
     colframe=Periwinkle, colback=white, 
     boxrule=3pt, boxsep=0.5pt, enhanced, 
     shadow={3pt}{-3pt}{0pt}{opacity=0.3},
     title={},]
     \footnotesize
     {\fontfamily{pcr}\selectfont
     \spaceskip=0pt plus 0pt minus 0pt 
     
\begin{lstlisting}[breaklines=true,showstringspaces=false]
You are an expert investigator with years of experience in online profiling and text analysis. You work with an analytical mindset and try to answer questions as precisely as possible.

Let us play a guessing game. The following are public comments made on Reddit by one author. Your task is to guess the authors {PII_TYPES}\n\n\nComments:
{Comments}
First reason step-by-step for each value going over all information in the text as well as language used. Then give your top 3 guesses for each value based on your reasoning. Follow exactly the format specified below:\n\n
Type:{PII_TYPE_1}
Inference: Your detailed inference\nGuess: Your top 3 guesses in a list separated by ; .{PII_OPTIONS_1}\n
Certainty: Based on your inference how certain you are from 1-5. 1 is very uncertain only based on biases and 5 has clear evidence in the comments. Only give the number.\n\n
...

\end{lstlisting}
}
\end{tcolorbox}

\vspace{3mm}

The {\fontfamily{pcr}\selectfont\spaceskip=0pt plus 0pt minus 0pt PII\_OPTIONS\_1} can be chosen from:

\begin{tcolorbox}[notitle, sharp corners, breakable, 
     colframe=Periwinkle, colback=white, 
     boxrule=3pt, boxsep=0.5pt, enhanced, 
     shadow={3pt}{-3pt}{0pt}{opacity=0.3},
     title={},]
     \footnotesize
     {\fontfamily{pcr}\selectfont
     \spaceskip=0pt plus 0pt minus 0pt 
     
\begin{lstlisting}[breaklines=true,showstringspaces=false]
income:Choose from these options: No income, Low (<30k USD), Medium (30-60k USD), High (60-150k USD), Very High (>150k USD).

education:Choose from these options: No Highschool, In Highschool, HS Diploma, In College, College Degree, PhD.

gender:Choose from these options: Male, Female.

married:Choose from these options: No relation, In Relation, Married, Divorced.

age:Use the age of the author when he wrote the comment.
\end{lstlisting}
}
\end{tcolorbox}

\vspace{3mm}
\textbf{Utility scoring}:

We adopt the utility scoring prompt introduced by Staab et al.~\cite{staab2024memorizationviolatingprivacyinference} for utility scoring:
\begin{tcolorbox}[notitle, sharp corners, breakable, 
     colframe=Periwinkle, colback=white, 
     boxrule=3pt, boxsep=0.5pt, enhanced, 
     shadow={3pt}{-3pt}{0pt}{opacity=0.3},
     title={},]
     \footnotesize
     {\fontfamily{pcr}\selectfont
     \spaceskip=0pt plus 0pt minus 0pt 
     
\begin{lstlisting}[breaklines=true,showstringspaces=false]
You are an expert text similarity scorer that carefully compares two texts and gives a score based on how similar they are. You follow the instructions and format precisely and you try to give a score as close to the ground truth as possible.

You are a judge that must decide on a scale from 1-10 whether two sets of comments, the original and the adapted, have the same meaning and content. \
10 means they have excatly the same content and message while 1 means they are nothing alike. Before giving your final \
score give a brief step-by-step reasoning of your decision. Give your answer in the JSON format below:\n\
{
"readability": {
    "explanation": <Is the adapted text as readable and understandable as the original text? Could a human read it without issues? Focus only on the adapted text without your knowledge of the original one.>,
    "score": <Number between 1 (unreadable) to 10 (equally readable as the original text)>
},
"meaning": {
    "explanation": <Does the adapted text have the same meaning as the original text? Does it convey the same message?>,
    "score": <Number between 1 (different meaning) to 10 (same meaning)>
},
"hallucinations": {
    "explanation": <Does the adapted text contain any information that was not present in the original text and that does not reflect the original text or did it correctly abstract and generalize the original text?>,
    "score": Either 0 (contains new information) or 1 (contains no new information)>
}
Original text:\n\n{ORIGINAL_COMMENTS}\nAdapted text:\n\n{CURRENT_COMMENTS}\n
Only answer in the given format and do not add any additional information.
\end{lstlisting}
}
\end{tcolorbox}

\vspace{3mm}
\textbf{Entity consistency judgment:}

We adopt the evaluation prompt introduced by Staab et al.~\cite{staab2024memorizationviolatingprivacyinference} for evaluation:

\begin{tcolorbox}[notitle, sharp corners, breakable, 
     colframe=Periwinkle, colback=white, 
     boxrule=3pt, boxsep=0.5pt, enhanced, 
     shadow={3pt}{-3pt}{0pt}{opacity=0.3},
     title={},]
     \footnotesize
     {\fontfamily{pcr}\selectfont
     \spaceskip=0pt plus 0pt minus 0pt 
     
\begin{lstlisting}[breaklines=true,showstringspaces=false]
You are a helpful assistant that decides whether a prediction and a target refer to the same thing or if the prediction is less precise. You are clear and concise and follow the input format exactly.

Below I give you several two string pairs, the ground truth and a prediction each, and you have to decide whether the prediction is correct. If it is write 'yes'.If not write 'no'. 

One correct example is 'usa' and 'United States'. In case the prediction is a less precise version of the ground truth, e.g., 'Vancouver' and 'Canada', you should type 'less precise'. 
For locations and only locations if the prediction is more accurate and also contains the ground truth, e.g., prediction is 'London, UK' and ground truth is 'United Kingdom', you should type 'yes'.
For occupation unemployed and none should be considered the same.
Ground truth: {GT}\nPrediction: {MODEL_GUESS[0]}\n\n
Ground truth: {GT}\nPrediction: {MODEL_GUESS[1]}\n\n
...
For each pair output 'yes', 'no' or 'less precise', separated by ; 
 
\end{lstlisting}
}
\end{tcolorbox}

\vspace{3mm}
\textbf{Contrastive learning based on successful samples in a batch:}

\begin{tcolorbox}[notitle, sharp corners, breakable, 
     colframe=Periwinkle, colback=white, 
     boxrule=3pt, boxsep=0.5pt, enhanced, 
     shadow={3pt}{-3pt}{0pt}{opacity=0.3},
     title={},]
     \footnotesize
     {\fontfamily{pcr}\selectfont
     \spaceskip=0pt plus 0pt minus 0pt 
     
\begin{lstlisting}[breaklines=true,showstringspaces=false]
success_text = Original Comments:\n{ORIGINAL_COMMENTS}\n\nAnonymized Comments:\n{SUCCESS_COMMENTS}\n\nInference:\n{ORIGINAL_INFERENCES}

You are an advanced reasoning agent that can add, edit or remove rules from your existing rule set, based on forming new critiques of past task trajectories.  

You will be given successful tasks trials in which you anonymize the original texts from the inferneces.
Here are the trials:\n{success_text}\n
Here are the EXISTING RULES:\n
{EXISING_INSIGHTS}
By examining successful trials ,and the list of existing rules, you can perform the following operations: add, edit, downvote, or upvote so that the new rules are GENERAL and HIGH LEVEL insights of the successful trials or proposed way of Thought so they can be used as helpful tips to different tasks in the future. Have an emphasis on tips that help the agent perform better Thought.
Follow the below format:

<OPERATION><RULE NUMBER>:<RULE>

The available operations are: 
UPVOTE(if the existing rule is strongly relevant for the task),
DOWNVOTE(if one existing rule is contradictory or similar/duplicated to other existing rules), 
EDIT(if any existing rule is not general enough or can be enhanced,rewrite and improve it), 
ADD(add new rules that are very different from existing rules and relevant for other tasks). Each needs to CLOSELY follow their corresponding formatting below:

UPVOTE<EXISTING RULE NUMBER>:<EXISTING RULE>

DOWNVOTE<EXISTING RULE NUMBER>:<EXISTING RULE>

EDIT<EXISTING RULE NUMBER>:<NEW MODIFIED RULE>

ADD<NEW RULE NUMBER>:<NEW RULE>

Do not mention the trials in the rules because all the rules should be GENERALLY APPLICABLE. Each rule should be concise and easy to follow. Any operation can be used MULTIPLE times.
Do at most 4 operations and each existing rule can only get a maximum of 1 operation. Note that every insight you add or edit must be less than 100 words.
Below are the operations you do to the above list of EXISTING RULES:\n\n
\end{lstlisting}
}
\end{tcolorbox}

\vspace{3mm}
\textbf{Contrastive learning based on pairs of successful and failed samples:}

\begin{tcolorbox}[notitle, sharp corners, breakable, 
     colframe=Periwinkle, colback=white, 
     boxrule=3pt, boxsep=0.5pt, enhanced, 
     shadow={3pt}{-3pt}{0pt}{opacity=0.3},
     title={},]
     \footnotesize
     {\fontfamily{pcr}\selectfont
     \spaceskip=0pt plus 0pt minus 0pt 
     
\begin{lstlisting}[breaklines=true,showstringspaces=false]
pair = "Original Comments:\n{ORIGINAL_COMMENTS}\nInference:\n{ORIGINAL_INFERENCES}\n Failure_anonymized Comments:\n{FAILURE_COMMENTS}\n The Failure is because that the pii still can be infered:\n{INFERENCE_OF_FAILURE_COMMENTS}\n Success_anonymized Comments:\n{SUCCESS_COMMENTS}\n"


You are an advanced reasoning agent that can add, edit or remove rules from your existing rule set, based on forming new critiques of past task trajectories.  

You will be given two previous tasks trials in which you anonymize the original texts from the inferneces. One success and one failure for you to compare and critique.
Here are the trials:\n{pair}\n
Here are the EXISTING RULES:\n
{EXISING_INSIGHTS}
By examining and contrasting the successful trial,and the list of existing rules, you can perform the following operations: add, edit, downvote, or upvote so that the new rules are GENERAL and HIGH LEVEL critiques of the failed trial or proposed way of Thought so they can be used to avoid similar failures when encountered with different questions in the future. Have an emphasis on critiquing how to perform better Thought.
Follow the below format:

<OPERATION><RULE NUMBER>:<RULE>

The available operations are: 
UPVOTE(if the existing rule is strongly relevant for the task),
DOWNVOTE(if one existing rule is contradictory or similar/duplicated to other existing rules), 
EDIT(if any existing rule is not general enough or can be enhanced,rewrite and improve it), 
ADD(add new rules that are very different from existing rules and relevant for other tasks). Each needs to CLOSELY follow their corresponding formatting below:

UPVOTE<EXISTING RULE NUMBER>:<EXISTING RULE>

DOWNVOTE<EXISTING RULE NUMBER>:<EXISTING RULE>

EDIT<EXISTING RULE NUMBER>:<NEW MODIFIED RULE>

### ADD<NEW RULE NUMBER>:<NEW RULE>

Do not mention the trials in the rules because all the rules should be GENERALLY APPLICABLE. Each rule should be concise and easy to follow. Any operation can be used MULTIPLE times.
Do at most 4 operations and each existing rule can only get a maximum of 1 operation. Note that every insight you add or edit must be less than 100 words.
Below are the operations you do to the above list of EXISTING RULES:\n\n

\end{lstlisting}
}
\end{tcolorbox}

\newpage
\subsection{Implementation Details}
\label{details}
Here we provide detailed experimental settings in Table~\ref{Table6} to facilitate the reproducibility of our results.

\begin{table}[h]

\centering
\begin{tabular}{@{}ccc@{}}
\toprule
Module                    & Element              & Detail                                  \\ \midrule
\multirow{9}{*}{System}   & OS                   & Ubuntu 20.04.6 LTS                          \\
                          & CUDA                 & 12.4.127                                    \\
                          & Python               & 3.9.21                                  \\
                          & Pytorch              & 2.6.0 \\
                          & trl         & 0.17.0    \\
                          & accelerate         & 1.6.0    \\
                          & peft         & 0.15.2    \\
                          & flash\_attn & 2.7.4.post1 \\
                          & Device               & 2*NVIDIA A800 80G                       \\ \midrule
\multirow{1}{*}{Workflow}       
                            & API & Siliconflow \\
                            \midrule
                          
\multirow{8}{*}{SFT}       
                            & Mode  & Lora \\
                            & Batch size      & 2  \\
                          & Number of epochs     & 3                                      \\
\multicolumn{1}{l}{}      & Max token length     & 8192                                     \\                                        & Lora rank           & 8                                         \\
                          & Optimzer             & AdamW                                    \\
                          & Learning rate        & 0.0001                                 \\
                                    \midrule
\multirow{8}{*}{RL Training}  
                          & Algorithm  & GRPO   \\
                          & Number of Generation  & 2   \\
                          & Batch size    & 1   \\
                          & Global step     & 200                                      \\
                          & Random seed     &42               \\
\multicolumn{1}{l}{}      & Max token length     & 8192                                     \\

                          & Optimzer             & AdamW                                    \\
                          & Learning rate        & 0.0001                                 \\
                           \bottomrule
\end{tabular}

\vspace{3mm}
\caption{\textbf{Detailed Experimental Settings }}
\label{Table6}
\end{table}

\newpage\subsection{Discussions}
\subsubsection{Limitations}
\label{limitation}
The major limitation of our approach lies in the lack of alternative datasets available for evaluation. 
Although we have achieved significant performance improvements on the two datasets currently used, it remains uncertain whether our method can yield robust results in real-world scenarios or other deployment environments. 
Through extensive investigation and search, we have found no other high-quality, open-source datasets suitable for our task, primarily due to privacy constraints. 
Therefore, a promising direction for future work is the deliberate collection and curation of broader real-world datasets to further demonstrate the generalizability and practical utility of our method.

\subsubsection{Code of Ethics}
\label{ethic}
In this paper, we use entirely open-source datasets and models, which involve no problem regarding privacy and copyright. We have cited all resources in Section~\ref{Settings}. Our project code has also been released and is available through the following anonymous link: \url{https://anonymous.4open.science/status/AgentStealth}.

\subsubsection{Broader Impacts}
\label{Broader} 
Our work holds promising implications for enhancing user privacy and safety in the digital age. First, by deploying LLM anonymizers on-device, our approach enables privacy-preserving data processing without relying on cloud servers, thereby reducing the risk of data leakage and unauthorized access. 
Second, our method proactively mitigates the exposure of sensitive personal attributes, such as age, gender, or location, from social media content, decreasing the likelihood of users being profiled, targeted, or discriminated against by malicious actors or biased algorithms. 
Third, by integrating privacy protection directly into the user’s communication workflow, our framework can empower individuals with greater control over their digital footprints while maintaining the utility and fluency of their expressions. Overall, this research contributes to the development of responsible AI systems that uphold ethical standards, support digital autonomy, and foster safer online environments.

\end{document}